\documentclass{article}
\usepackage{ijcai21}

\usepackage{times}

\usepackage{soul}
\usepackage{url}
\usepackage[hidelinks]{hyperref}
\usepackage[utf8]{inputenc}
\usepackage[small]{caption}
\usepackage{graphicx}
\usepackage{amsmath}
\DeclareMathOperator*{\argmax}{argmax} 
\usepackage{booktabs}
\title{Pseudo-labelling Enhanced Media Bias Detection}

\author{
Qin Ruan$^1$\footnote{Contact Author}\and
Brian Mac Namee$^2$\and
Ruihai Dong$^2$\\
\affiliations
$^1$School of Computer Science, University College Dublin, Dublin, Ireland\\
$^2$Insight Centre for Data Analytics, University College Dublin, Dublin, Ireland\\
\emails
qin.ruan@ucdconnect.ie,
\{brian.macnamee, ruihai.dong\}@ucd.ie,
}

\begin{document}
\maketitle

\begin{abstract}
Leveraging unlabelled data through weak or distant supervision is a compelling approach to developing more effective text classification models. This paper proposes a simple but effective data augmentation method, which leverages the idea of pseudo-labelling to select samples from noisy distant supervision annotation datasets. The result shows that the proposed method improves the accuracy of biased news detection models.

\end{abstract}

\section{Introduction}


The mission of news media is to convey accurate and fair opinions to readers. However, the reality of modern media often fails to live up to this ideal. Bias exists in news reporting, and sometimes this is as a subtle as only giving a one-sided version of a story \cite{kiesel2019semeval}. The side effects of media bias---for example, the potential to distort the reader's view of the world---have been widely recognised \cite{bernhardt2008political}. How to automatically detect media bias quickly and accurately, so as to make users aware of the bias in the media they consume and provide a fairer environment for news reporting, has been a long-term challenge.


For handling this challenge, important datasets, which are the prerequisite for training machine learning algorithms to accurately recognize biased news articles, have been created. Mainly, these datasets can be divided into two categories based on how the datasets have been annotated: \emph{distant supervision} and \emph{manual annotation}. In the  context of news media distant supervision refers to automatically labelling characteristics such as the  political ideology of news articles not based on detailed analysis of the news articles themselves, but rather based on the publishers' characteristics. This allows quick and inexpensive production of a large amount of labelled data, but with significant amounts of noise and errors. Compared with distant supervision, manual labelling by experts or crowdsourcing workers provides better quality, but this method is time-consuming and very expensive ~\cite{cremisini2019challenging}. 

Kiesel et al ~\shortcite{kiesel2019semeval} describe an important dataset that includes annotations of both of these types. A large news bias dataset for detecting hyperpartisan (biased/unbiased) articles, including 754,000 articles, was automatically annotated by distant supervision based on the articles' publisher. A smaller dataset containing 1,273 instances was manually labelled. It is worth noting that the best models that have achieved the best performance using these datasets are trained using only the manually labelled dataset. This implies that the use of the data annotated using distant supervision, \emph{by-publisher} data, is challenging and may lead to poorly performing models. 

In this paper, we propose an overlap-checking mechanism that uses the idea of pseudo-labelling to select samples from the noisy distant supervision dataset, then jointly trains a model on manually labelled data and selected distant supervision data. The evaluation result on the hyperpartisan news detection task shows that the mechanism proposed is effective.

\section{Overlap-checking Mechanism}

\begin{figure}
    \centering
    \includegraphics[scale=0.42]{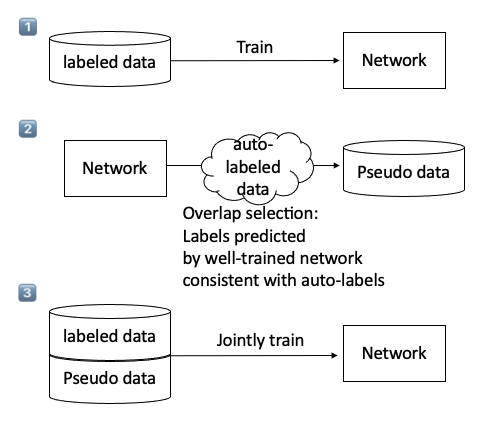}
    \caption{An overview of the overlap-checking mechanism}
    \label{fig:overlap-checking framework}
\end{figure}

An overview of the proposed overlap-checking mechanism is presented in Figure \ref{fig:overlap-checking framework}. The framework contains three steps. The network first trains on manually labelled data until it converges; then the training leverages the overlap-checking mechanism to select a batch of the pseudo-labelled data; finally, the model is re-train jointly with labelled data and pseudo-labelled data.

\vspace{3px}
\noindent
\textbf{Pseudo-labeling method}: For unlabelled samples, the probability distribution of the model prediction is used as an indicator to pseudo-label the data ~\cite{lee2013pseudo}. The method belongs to the branch of semi-supervised Learning. Using this simple and efficient method, the system can easily add more data to help re-train the model. ~\cite{he2017unified} proved that using a batch of samples with the highest prediction probability of the model can help enhance the performance of the model.

\vspace{3px}
\noindent
\textbf{Overlap-checking mechanism}: The vanilla pseudo-labelling method selects the class with the highest predicted probability from the completely unlabelled dataset as the pseudo label of the sample. Assuming that there are $L$ categories, denote $l$ as category instance, where a value of $1$ represents category $l$ is selected and $0$ not selected, the formula is as follows:

$$y^{'}=
\begin{cases}
1& \text{if }l=\argmax_{l \in L}f(x)^{'}\\
0& \text{otherwise}
\end{cases}$$


\vspace{3px}
\noindent
\textbf{Label combination}: We combine both the pseudo-labelled annotation and the distant supervision annotation on the by-publisher dataset by considering their consistency. We denote the distant supervision dataset as $A$, the pseudo-labelled dataset as $P$, and the intersection set of $A$ and $P$ as candidate set $C = A \cap P$.

Eventually, the top $N$ pseudo samples are returned on the basis of descending order of the predicted probability value, where $N$ represents the expected number of pseudo-samples.

\section{Experiments}
This section describes an evaluation of the performance of the overlap-checking mechanism on the hyperpartisian news detection task.

\vspace{3px}
\noindent
\textbf{Hyperpartisan news detection}: Task 4  of the 2019 SemEval challenge  ~\cite{kiesel2019semeval} focused on recognizing biased news articles. the related dataset contains 1,273 manually labelled instances and 754,000 instances annotated based on the characteristic of publishers. With the aim of fairness, the test data is not publicly accessible and  competitors submit executable scripts for evaluation.

\vspace{3px}
\noindent
\textbf{Training details}: We train the detector from the best team ~\cite{jiang2019team} of the competition, which builds an Elmo-based sentence encoder to encode sentences to high-dimensional semantic vectors. These vectors are passed into different initialized convolutional neural layers and batch normalization layers in parallel. The final output is a dense layer followed by a sigmoid function which concatenates the output of previous layers.

Our experimental setting follows the same configuration on the network side. Different from the best team scheme that only uses manually labelled data, we also select a percentage of data from the noisy by-publishers by adopting our proposed overlap-checking mechanism.

\vspace{3px}
\noindent
\textbf{Results}: Following the custom of the hyperpartisian news detection challenge we name our approach Otto Chriek, after a fictional journalistic character. Table \ref{tab:Table.learderboard} compares the performance of our approach that integrates the overlap-checking mechanism with the top 4  existing results on the challenge leaderboard, indicating that our approach surpasses all existing models in accuracy, recall and f1 score.

\begin{table}[!tp]
    \begin{tabular}{c|c|c|c|c}
    \toprule
    \textbf{Submission} & \multicolumn{4}{|c}{\textbf{By-article Dataset}} \\
    \hline
    Team name & Acc. & Prec. & Recall & F1 \\
    \midrule
    Tom Jumbo Grumbo & 0.806 & 0.858 & 0.732 & 0.790 \\
    Sally Smedley & 0.809 & 0.823 & 0.787 & 0.805 \\
    Vernon Fenwick & 0.820 & 0.815 & 0.828 & 0.821 \\
    Bertha von Suttner & 0.822 & \textbf{0.871} & 0.755 & 0.809 \\
    \midrule
    Otto Cheirk & \textbf{0.831} & 0.823 & \textbf{0.844} & \textbf{0.834} \\
    \bottomrule
    \end{tabular}
    \caption{The leaderboard scores of Hyperpartisian News Detection}
    \label{tab:Table.learderboard}
\end{table}

\section{Conclusion}

The paper demonstrates a simple overlap checking mechanism. On the dataset labelled by distant supervision, a well-trained model is used to output the model prediction labels, and then the dual check is used to select data from a noisy auto-annotation dataset. Our experiments on biased news article detection show that our proposed method could reduce the risk of using a noisy dataset and help improve the model performance.

\section{Acknowledgement}
This publication has emanated from research conducted with the financial
support of Science Foundation Ireland under Grant number 18/CRT/6183. For the purpose
of Open Access, the author has applied a CC BY public copyright licence to any
Author Accepted Manuscript version arising from this submission.

\bibliographystyle{named}
\bibliography{main}

\end{document}